\documentclass[letterpaper]{article} % DO NOT CHANGE THIS
\usepackage{aaai2026}  % DO NOT CHANGE THIS
\usepackage{times}  % DO NOT CHANGE THIS
\usepackage{helvet}  % DO NOT CHANGE THIS
\usepackage{courier}  % DO NOT CHANGE THIS
\usepackage[hyphens]{url}  % DO NOT CHANGE THIS
\usepackage{graphicx} % DO NOT CHANGE THIS
\usepackage{natbib}  % DO NOT CHANGE THIS AND DO NOT ADD ANY OPTIONS TO IT
\usepackage{caption} % DO NOT CHANGE THIS AND DO NOT ADD ANY OPTIONS TO IT
\usepackage{algorithm}
\usepackage{algorithmic}
\usepackage{amsmath} 
\usepackage{amssymb}
\usepackage{amsfonts}
\usepackage{subcaption}
\usepackage{multirow}
\usepackage{booktabs}

\usepackage{newfloat}
\usepackage{listings}
\DeclareCaptionStyle{ruled}{labelfont=normalfont,labelsep=colon,strut=off} % DO NOT CHANGE THIS
\floatstyle{ruled}
\newfloat{listing}{tb}{lst}{}
\floatname{listing}{Listing}
\title{CO-RFT: Efficient Fine-Tuning of Vision-Language-Action Models through Chunked Offline Reinforcement Learning}
\author{
    Dongchi Huang\textsuperscript{1}\equalcontrib,
    Zhirui Fang\textsuperscript{2}\equalcontrib,
    Tianle Zhang\textsuperscript{3},
    Yihang Li\textsuperscript{3},
    Lin Zhao\textsuperscript{3},
    Chunhe Xia\textsuperscript{1}\thanks{Correspondence to: Chunhe Xia at xch@buaa.edu.cn.}
}
\affiliations{
    \textsuperscript{1}School of Computer Science and Engineering, Beihang University \\
    \textsuperscript{2}Tsinghua Shenzhen International Graduate School, Tsinghua University \\
    \textsuperscript{3}JD Explore Academy
}

\usepackage{bibentry}
\begin{document}

\maketitle

\begin{abstract}
Vision-Language-Action (VLA) models demonstrate significant potential for developing generalized policies in real-world robotic control. This progress inspires researchers to explore fine-tuning these models with Reinforcement Learning (RL). However, fine-tuning VLA models with RL still faces challenges related to sample efficiency, compatibility with action chunking, and training stability. To address these challenges, we explore the fine-tuning of VLA models through offline reinforcement learning incorporating action chunking.
In this work, we propose \textbf{Chunked RL}, a novel reinforcement learning framework specifically designed for VLA models. Within this framework, we extend temporal difference (TD) learning to incorporate action chunking, a prominent characteristic of VLA models. Building upon this framework, we propose \textbf{CO-RFT}, an algorithm aimed at fine-tuning VLA models using a limited set of demonstrations (30 to 60 samples). Specifically, we first conduct imitation learning (IL) with full parameter fine-tuning to initialize both the backbone and the policy. Subsequently, we implement offline RL with action chunking to optimize the pretrained policy.
Our empirical results in real-world environments demonstrate that \textbf{CO-RFT} outperforms previous supervised methods, achieving a 57\% improvement in success rate and a 22.3\% reduction in cycle time. Moreover, our method exhibits robust positional generalization capabilities, attaining a success rate of 44.3\% in previously unseen positions.

\end{abstract}

\section{Introduction}
Vision-Language-Action (VLA) models, which integrate perception and language understanding for embodied control, have emerged with promising potential to develop general policies for real-world robotic control. These models are based on large vision-language models that are pretrained on internet-scale data and subsequently trained using extensive, heterogeneous robot demonstration datasets. This paradigm has achieved significant success, demonstrating capabilities in mastering a variety of tasks.
However, fine-tuning methods for VLA models primarily rely on supervised fine-tuning (SFT) through behavioral cloning of demonstrations. This process faces significant challenges: the model's performance is heavily reliant on the quality and quantity of task-specific data, and it struggles to generalize to out-of-distribution (OOD) scenarios.

Reinforcement Learning (RL) has the potential to address the challenges currently faced by SFT, leading to a more promising solution for fine-tuning VLA models. RL, which directly optimizes the policy through trial and error, possesses the ability to surpass the constraints of expert data, enabling the learning of corrective behaviors and the generalization to previously unseen scenarios. Meanwhile, recent breakthroughs in applying RL to large language models (LLMs) have demonstrated that RL can result in substantially better OOD performance compared to SFT. Moreover, there is evidence that offline RL can learn a better policy than behavior cloning, owing to the capability of Q-learning to prioritize critical decisions. Drawing from these encouraging results in previous studies, extensive research has been conducted to explore the fine-tuning of VLA models using RL, yielding impressive performance outcomes. Including offline RL \cite{zhang2025reinbot}, online RL \cite{chen2025conrftreinforcedfinetuningmethod, shu2025rftfreinforcementfinetuningembodied, tan2025interactiveposttrainingvisionlanguageactionmodels, guo2025improvingvisionlanguageactionmodelonline, zhang2024grape}
and test-time RL \cite{nakamoto2024steering, song2025hume, du2024errroboticrapidvaluebased}.

Despite these promising achievements, applying RL to fine-tuning VLA models in real-world scenarios still faces significant challenges. For online RL, it is essential to establish the infrastructure necessary to ensure both autonomous learning and training stability for large models. However, existing methods are predominantly confined to simulators; those trained in real-world settings either limit themselves to small-scale VLA models \cite{chen2025conrftreinforcedfinetuningmethod} or circumvent this issue by iterating between RL and IL \cite{guo2025improvingvisionlanguageactionmodelonline}.
For test-time RL \cite{nakamoto2024steering, song2025hume, du2024errroboticrapidvaluebased}, this approach samples multiple actions from the policy, ranks them using the value function, and selects the top actions for execution. Due to this characteristic, such methods can yield only marginal performance improvements while extending inference time. 
In contrast to the paradigms discussed above, we contend that offline RL offers a more viable solution for fine-tuning VLA models. Offline RL optimizes the policy on static datasets, thereby eliminating the reliance on the infrastructural efforts required for online RL. Otherwise, offline RL directly optimizes the policy without compromising inference efficiency and can yield superior policies compared to test-time RL. 

Inspired by these insights, we investigate the efficient fine-tuning of VLA models through offline RL that incorporates action chunking. Action chunking, a prevalent feature in VLA models, enhances action smoothness and mitigates non-Markovian behavior; however, it has been overlooked in recent research. 
To address this issue, we propose \textbf{Chunked RL}, a novel reinforcement learning framework specifically designed for VLA models. 
Specifically we implement the critic network using a transformer block. It receives as input the current state and an action chunk $(s_t, {a_t, a_{t+1}, \cdots, a_{t+h-1}})$, and subsequently predicts a sequence of Q-values for the action chunk. Each Q-value is responsible for predicting the K-step returns for all $K=1\dots N$. We then utilize the mean of all Q-values to optimize the action chunk. Previous studies demonstrate that incorporating action chunks in reinforcement learning can enhance sample efficiency, stability, and the ability to handle sparse reward settings \cite{tian2025chunkingcritictransformerbasedsoft, li2025toperl}, which is particular suitable for fine-tuning VLA models. 
Building upon this framework, we propose \textbf{CO-RFT}, a two-stage RL algorithm designed to transfer VLA models to the current workspace and enhance their performance in downstream tasks. In the first stage, we perform imitation learning (IL) with full parameter fine-tuning to transfer the VLM backbone to the current workspace and retrain the policy head to accommodate the current embodiment. In the second stage, we implement offline RL with action chunking to further enhance performance.
Our empirical results in real-world environments demonstrate that \textbf{CO-RFT} outperforms SFT, achieving a 57\% improvement in success rate and a 22.3\% reduction in cycle time. Moreover, our method exhibits robust positional generalization capabilities, attaining a success rate of 44.3\% in previously unseen positions.

\begin{figure*}[!htb]
    \centering
    \includegraphics[width=1.0\linewidth]{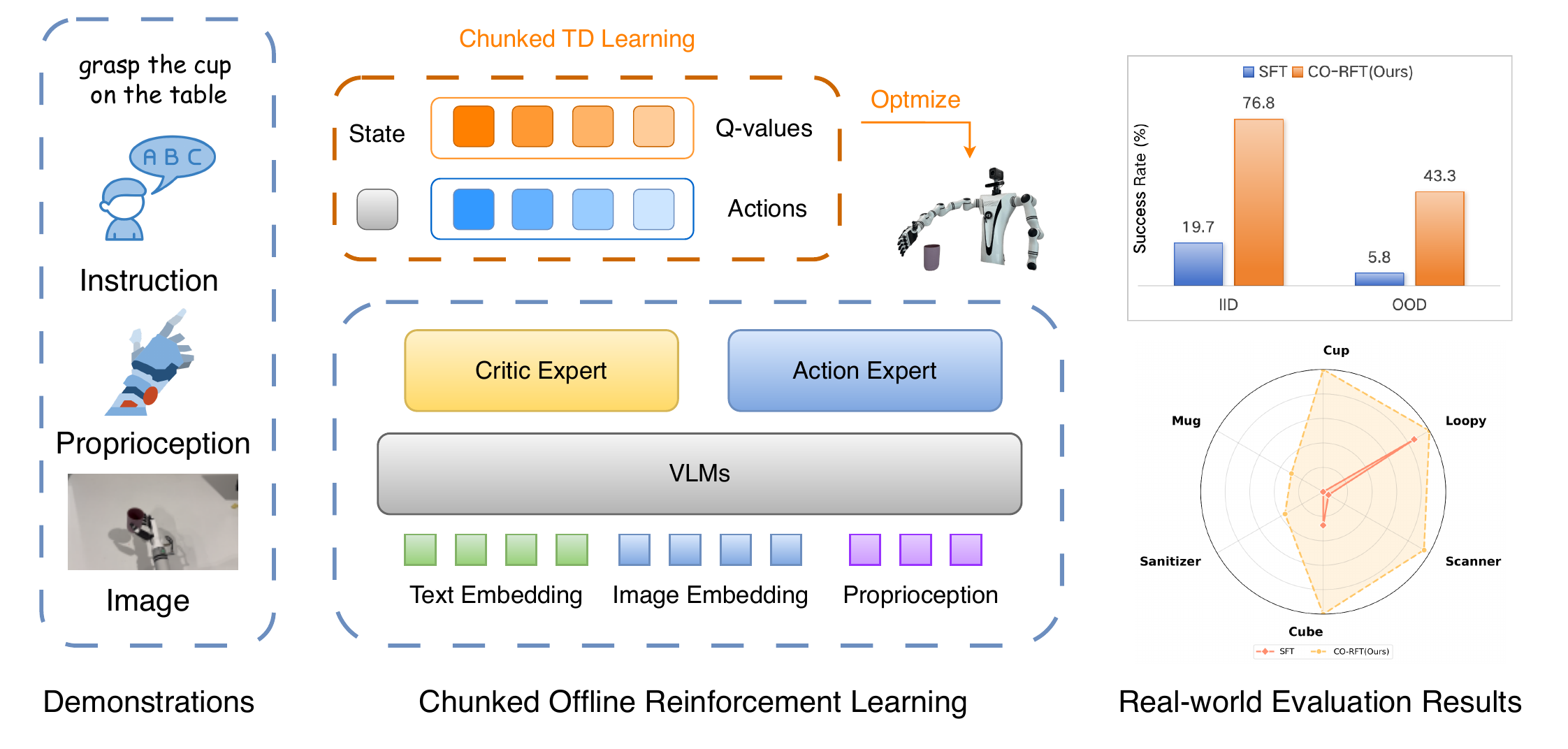}
    \caption{Overview of CO-RFT. The demonstrations consists of language instructions, proprioception and images. In our framework, in addition to VLMs and action experts commonly utilized in VLA models, we incorporate a critic expert to generate Q-values. Unlike standard RL, the critic expert predicts a sequence of Q-values based on the action chunk.}
    \label{fig:overview}
\end{figure*}
.

% \citet{zhang2025reinbot} achieves a deeper understanding of the data quality distribution by predicting dense returns with offline RL. 

% Offline RL \cite{qtransformer, springenberg2024pac, song2024germ, zhao2025moreunlockingscalabilityreinforcement} 

\section{Related Work}
\paragraph{Vision-Language-Action Models.} 
VLA models demonstrate remarkable potential for developing generalizable robot policies through imitation learning on extensive robot datasets across diverse tasks and embodiments.
\cite{kim2024openvlaopensourcevisionlanguageactionmodel, contributors2025agibotworld,li2024generalistrobotpoliciesmatters, nvidia2025gr00tn1openfoundation, zhang2025hirtenhancingroboticcontrol, shi2025hirobotopenendedinstruction, black2024pi0visionlanguageactionflowmodel, pertsch2025fastefficientactiontokenization, qu2025spatialvlaexploringspatialrepresentations, wu2023unleashinglargescalevideogenerative, cheang2024gr2generativevideolanguageactionmodel}.
These models primarily utilize a high-capacity Vision-Language Model (VLM) as a backbone to process language instructions and visual inputs, while integrating an action expert to derive robotic actions, demonstrating an impressive capability to master diverse tasks. However, these Vision-Language Action (VLA) models trained through imitation learning still face challenges related to the significant costs of collecting expert demonstrations, low inference speed, and their inability to effectively address out-of-distribution (OOD) scenarios.

\paragraph{Reinforcement Learning for VLA models.}
Inspired by the promising success that RL has achieved in fine-tuning LLMs and VLMs, recent efforts have focused on integrating RL to improve the performance of VLA models.
ReinboT  \cite{zhang2025reinbot} integrates RL returns maximization to enhance robotic manipulation capabilities, exhibiting superior few-shot learning and out-of-distribution generalization capabilities in real-world tasks.
\citet{liu2025rlbringvlageneralization} formulate General Robotic Manipulation as multi-turn Conversation problems and fine-tune a robotic process reward model for reward densification, and employ Proximal Policy Optimization (PPO) for stable policy optimization.
\cite{zhang2024grape} propose trajectory-wise preference optimization (TPO) to fine-tuning VLA models  from the perspective of preference alignment.
\cite{hu2024flare} explore fine-tuning VLA models with large-scale RL, providing significant techniques for stable RL fine-tuning. 
\cite{liu2025rlbringvlageneralization} conduct an empirical study to investigate how RL fine-tuning affects the generalization ability of VLA models, identifying PPO as a more effective RL algorithm for fine-tuning VLA models.
iRe-VLA \cite{guo2025improvingvisionlanguageactionmodelonline} introduces a framework that effectively improves VLA models by iterating between online reinforcement learning and behavior cloning.
% This method exploits reinforcement learning solely to acquire high-quality trajectories, failing to fully leverage the self-improving nature of reinforcement learning.
ConRFT \cite{chen2025conrftreinforcedfinetuningmethod} presents a two-stage reinforcement learning framework that utilizes offline RL to initialize the value function, subsequently enhancing VLA models through online RL with human intervention \cite{luo2024hilserl}. 
% Although this method yields promising results, it is implemented on Octo-Small \cite{octo_2023}, which has only 27M parameters—a size that may be too small to reflect the generalization ability of VLA models.
Moreover, none of these methods address the compatibility issues of reinforcement learning and action chunking \cite{zhao2023learningfinegrainedbimanualmanipulation}, a common technique utilized by VLA models.

\paragraph{Action Chunking.}
Action chunking, wherein the policy predicts a sequence of actions at each decision step, has been demonstrated to enhance action smoothness and effectively address non-Markovian behavior in imitation learning (IL). Consequently, the notable success of action chunking in IL has inspired researchers to incorporate chunking into reinforcement learning (RL). \citet{tian2025chunkingcritictransformerbasedsoft} leverages the Transformer as a critic to predict the value of action chunks by integrating $n$-step returns with a transformer, thereby demonstrating sample efficiency, stability, and a strong capability to manage sparse rewards. Similarly, \citet{li2025toperl} integrate the Transformer-based critic with motion primitives in an online episodic RL setting, resulting in more stable and effective value learning. Additionally, \citet{seo2025reinforcement} introduce action chunking into the CQN, demonstrating sample efficiency and impressive performance in long-horizon and sparse reward tasks. Instead of learning the value function for each action within the action sequences, \citet{li2025reinforcementlearningactionchunking} estimate the value function for executing the entire action sequence. This design enhances sample efficiency and mitigates the exploration challenge due to the temporally coherent actions. These studies have demonstrated that incorporating action chunking in reinforcement learning (RL) can improve sample efficiency, address long-horizon and sparse reward tasks, and mitigate exploration challenges. These factors represent the primary obstacles encountered by VLA reinforced fine-tuning, which previous methods have neglected. Consequently, in this paper, we aim to propose a fine-tuning algorithm for VLA models leveraging reinforcement learning with action chunking.

\section{Preliminaries}
%=================================================================
% 
%                         MDP definition
% 
%=================================================================
\paragraph{Markov decision process (MDP).} RL learns policies that maximize cumulative rewards in a given environment, modeled as an MDP. Formally, we consider an MDP defined by a tuple $(\mathcal{S}, \mathcal{A}, P, r, \gamma)$, where both state $\mathcal{S}$ and action spaces $\mathcal{A}$ are continuous. Here, $P(s'|s,a)$ denotes the state transition probability, $r(s,a)$ is the reward function, and $\gamma \in [0,1]$ is the discount factor. The goal of RL is to find a policy $\pi(a|s)$ that maximizes the expected \textit{return}, which is the sum of discounted future rewards as $G_t = \sum_{i=0}^{\infty} \gamma^i r_{t+i}.$.
% $G_t(s_t, a_t) = \sum_{i=0}^{\infty} \gamma^i r_{t+i}$. 
% \begin{equation}
%     G_t = \sum_{i=0}^{\infty} \gamma^i r_{t+i}.
% \end{equation}

%=================================================================
% 
%                         Offline RL 
% 
%=================================================================

\paragraph{Offline RL.}
Given access to an offline dataset $\mathcal{D} = \{(s, a, r, s^\prime)\}$ collected using a behavior policy $\pi_\beta$, we aim to train a good policy $\pi$ and value function $Q_\theta(s, a)$ using the offline dataset $\mathcal{D}$.
Offline RL algorithms based on this basic recipe suffer from action distribution shift during training, because the target values for Bellman backups in policy evaluation use actions sampled from the learned policy $\pi$, but the Q-function $Q_\theta(s, a)$ is trained only on actions sampled from the behavior policy $\pi_\beta$ that produced the dataset $\mathcal{D}$. Since $\pi$ is trained to maximize Q-values, it may be biased towards out-of-distribution (OOD) actions with erroneously high Q-values. In standard RL, such errors can be corrected by attempting an action in the environment and observing its actual value. However, the inability to interact with the environment makes it challenging to deal with Q-values for OOD actions in offline RL. Typical offline RL methods mitigate this problem by constraining the learned policy $\pi$ away from OOD actions. 
% Note that Q-function training in offline RL does not suffer from state distribution shift, as the Bellman backup never queries the Q-function on out-of-distribution states. However, the policy may suffer from state distribution shift at test time.
CalQL \cite{nakamoto2024calqlcalibratedofflinerl} imposes an additional regularizer that penalizes the learned Q-function on out-of-distribution (OOD) actions while compensating for this pessimism on actions seen within the training dataset. The training objective of CalQL is given by:
\begin{equation}
    \label{eq:cql_training}
    \min_\theta { \alpha \mathcal{R}(\theta)
    + \frac{1}{2} {\mathbb{E}_{s, a, s^\prime\sim \mathcal{D}}\left[\left(Q_\theta(s, a) - B^{\pi}\overline{Q}(s, a)\right)^2 \right]}},
\end{equation}
where $B^{\pi}\overline{Q}(s, a)$ is the backup operator applied to the delayed target Q-network $\overline{Q}$, defined as $B^{\pi}\overline{Q}(s, a) := r(s, a) + \gamma \mathbb{E}_{a^\prime \sim \pi(a^\prime|s^\prime)}[\overline{Q}(s^\prime, a^\prime)]$. The second term is the standard TD error. The first term  $\mathcal{R}(\theta)$ is a conservative regularizer that aims to prevent overestimation in the Q-values for OOD actions by minimizing the Q-values under the policy $\pi$, and counterbalances by maximizing the Q-values of the actions in the dataset following the behavior policy $\pi_\beta$. The conservative regularizer $\mathcal{R}(\theta)$ of CalQL is given by:
\begin{equation}
\label{eq:calql}
\mathbb{E}_{s \sim \mathcal{D}, a \sim \pi} {[\max \left( Q_\theta(s,a), V^\mu(s) \right)]}
- \mathbb{E}_{s, a \sim \mathcal{D}}\left[Q_\theta(s,a)\right],
\end{equation}
this term was proposed to ensure that the learned Q-values closely match the range of ground-truth Q-values, thereby preventing unlearning during the online fine-tuning stage.
%=================================================================
% 
%                         N-step return 
% 
%=================================================================
\paragraph{The N-step return} The N-step return enhances the single-step TD return by integrating multiple future time-steps into the target. Unlike bootstrapping after a single time-step, the N-step return aggregates rewards over $N$ steps before employing the current value estimate for bootstrapping. These estimates are generally less biased than the 1-step return; however, they also exhibit greater variance. The N-step return is updated as follows:
% \begin{equation}
%     G_t^{(N)}(s_t, a_t) = \sum_{i=0}^{N-1} \gamma^i r_{t+i} + \gamma^N Q^\pi(s_{t+N}, a_{t+N}).
% \end{equation}
\begin{equation}
    G_t^{(N)}(s_t, a_t) = \sum_{i=0}^{N-1} \gamma^i r_{t+i} + \gamma^N Q^\pi(s_{t+N}, a_{t+N}),
    \label{eq:n_step_return_old}
\end{equation}
 As the effective horizon $N$ goes up, the value can propagates $N$ step backward (from $s_{t+N}$ to $s_t$). Consequently, the value estimate of $Q(s_t, a_t)$ allows for a $n$ times speed-up in terms of horizon $N$. Due to this benefit of $n$-step return, it has been commonly adopted in large-scale RL systems~\citep{mnih2016asynchronousmethodsdeepreinforcement, hessel2017rainbowcombiningimprovementsdeep, kapturowski2018recurrent, wurman2022outracing}.

\section{Methodology}
In this section, we first outline the two primary design principles of out method: (1) Q-learning on an action chunk, and (2) two-stage reinforced fine-tuning framework for VLA models. Finally, we describe practical implementations of our reinforced fine-tuning framework.
\subsection{Chunked Reinforcement Learning}
We aim to apply $Q$-learning on the temporally extended action space to be compatible with the action chunk technique in VLA models. We use $a_{t:t+h}$to denote a concatenation of $h$ consecutive actions: $\begin{bmatrix} a_t & \cdots & a_{t+h-1} \end{bmatrix} \in \mathbb{R}^{Ah}$ for notation convenience. Consequently, we formally design our actor-critic model as follows:
\begin{align*}
 &\text{\emph{Policy: }} \pi_\psi(a_{t:t+h} | s_t) :=\pi_\psi( a_t, a_{t+1}, \cdots, a_{t+h-1} | s_t) \\
  &\text{\emph{Critic: }} Q_\theta(s_t, { a_{t:t+h}}) := Q_\theta(s_t, {a_t, a_{t+1}, \cdots, a_{t+h-1}})  
\end{align*}

\paragraph{Chunked TD-learning}
Unlike normal 1-step TD-learning methods, which train a Q-function $Q(s_t, a_t)$ and a policy $\pi(a_t | s_t)$.
\begin{align}
    G_t(s_t, a_{t}) & \leftarrow r_t + \gamma Q(s_{t+1}, a_{t+1}).
\end{align}
we instead train both the critic and the actor with a span of $h$ consecutive actions $a_{t:t+h}$. In practice, this involves updating the critic and the actor on batches of transitions consisting of a state $s_t$, an action sequence $a_{t:t+h}$ and the state $h$ steps into the future, $s_{t+h}$. 
\begin{align}
    G_t^{(h)}(s_t, a_{t:t+h}) &\leftarrow \sum_{t'=t}^{t+h-1} \left[\gamma^{t'-t}r_{t'}\right] + \gamma^h Q(s_{t+h}, { a_{t+h:t+2h}}).
\end{align}
with ${a_{t+h:t+2h}} \sim \pi_\psi(\cdot | s_{t+h})$, and $\bar \theta$ being the target network parameters that are often an exponential moving average of $\theta$~\citep{mnih2013playingatarideepreinforcement}.
This value estimate of $Q(s_t, a_t)$ allows for a $n$ times speed-up in terms of the number of time steps that the value can propagate back across.

\paragraph{Chunked Critic Training Objective}
Given the next n actions, our critic is tasked to output all N-step returns (from 1 to $N$) for this action sequence. The critic is trained by minimizing the mean squared error (MSE) between its Q-value estimates and the corresponding N-step returns: Specifically, we train $Q_\theta$ with the following TD loss,
\begin{align}
    L(\theta) = \frac{1}{Nh} \sum_{k=1}^{N} \sum_{i=1}^{h} \left(Q_\theta(s_t, a_{t:t+i}) - G_t^{(h)}(s_t, a_{t:t+i})\right)^2
\end{align}

\begin{figure}[!htb]
    \centering
    \includegraphics[width=1.0\linewidth]{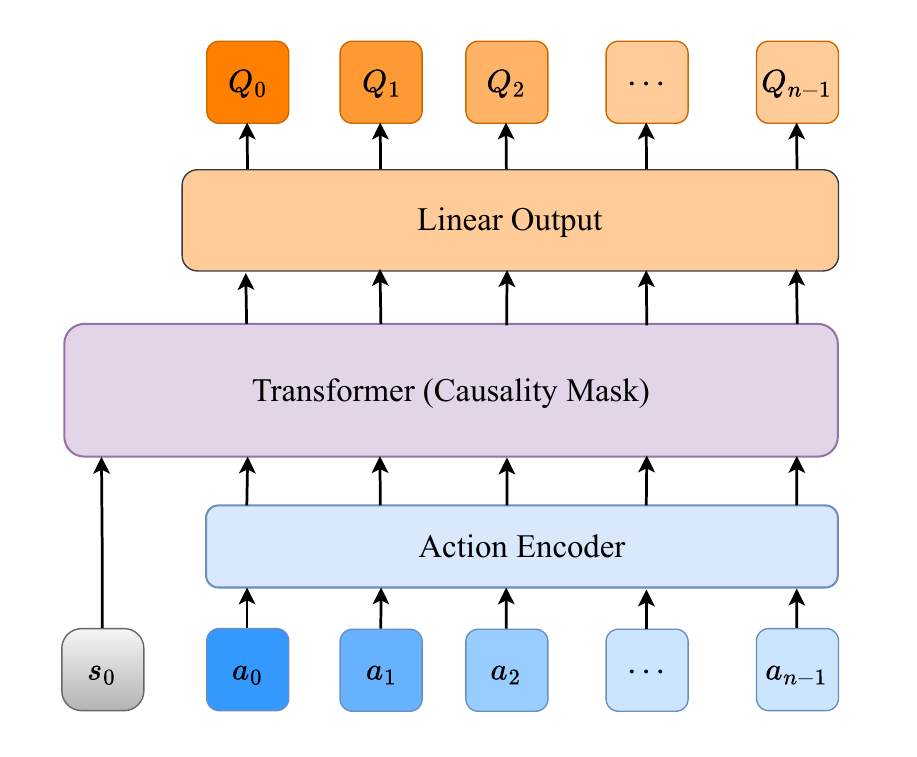}
    \caption{Visualization of Critic Network.}
    \label{fig:critic_network}
\end{figure}

\paragraph{Chunked Critic Network}
Chunk RL requires the learning of $H$ Q-values and, correspondingly, $H$ networks for this purpose. However, the associated computational cost is prohibitively high. To simplify this process, as illustrated in Figure.\ref{fig:critic_network}, we employ self-attention and a causal mask, utilizing state and action chunks as inputs. Among the $H+1$ embedding outputs generated, the last $H$ outputs correspond to the embeddings. Subsequently, the corresponding Q-values are derived from these embeddings. This design necessitates only a single network to learn all Q-values.

\subsection{Chunked Offline Reinforced Fine-tuning}
\textbf{CO-RFT} comprises two training stages. In the first stage, we perform Behavior Cloning (BC) with full parameter fine-tuning to initialize both the backbone and the policy. Subsequently, we employ offline RL with action chunking to optimize the pretrained policy.
\paragraph{Stage 1.BC}
Because existing VLA architectures are trained with gripper datasets and gripper robots, we must first transfer the VLM backbone and action head to our workspace and robot. Specifically, we collected 30 demonstrations via teleoperation and employed BC with the following objective:
\begin{equation}
    \mathcal{L}^{BC}=\mathbb{E}_{(s^e,a^e)\thicksim\mathcal{\tau}^e}\|a^e-\pi^{BC}(s^e)\|^2
\end{equation}
where, $\mathcal{\tau}^e$ represents expert demonstrations, $s^e$ and $a^e$ denote the corresponding state and action in expert demonstrations, and $\pi^{BC}$ signifies the learned behavior cloning policy. 

\paragraph{Stage 2.Chunked Offline RL} 
We aim to investigate methods for achieving superior performance compared to behavior cloning while utilizing the same dataset within the framework of offline RL. This approach enables more effective utilization of collected datasets while circumventing the stability, safety, and engineering challenges typically associated with online RL. 
Following equation.\ref{eq:cql_training} and equation.\ref{eq:calql}, we define the critic training object of chunked-CalQL:
\begin{align}
    \label{eq:chunked_calql_training}
    L(\theta) =
    &\mathbb{E}_{s, a, s^\prime\sim \mathcal{D}}\left[\left(Q_\theta(s, a_{t:t+i}) - B^{\pi}\overline{Q}(s, a_{t:t+i})\right)^2 \right] \nonumber \\
    +& \alpha ( \mathbb{E}_{s \sim \mathcal{D}, a \sim \pi} {[\max \left( Q_\theta(s,a_{t:t+i}), V^\mu(s) \right)]} \\
    -& \mathbb{E}_{s, a \sim \mathcal{D}}\left[Q_\theta(s,a_{t:t+i})\right]), \nonumber
\end{align}

\begin{align}
    L(\psi) = - \frac{1}{Nh} \sum_{k=1}^{N} \sum_{i=1}^{h} Q_\theta(s_t, a_{t:t+i})
\end{align}

\paragraph{Why CalQL?} However, challenges related to sparse rewards and sample efficiency persist in offline RL. To mitigate these challenges, we employ Chunked RL. Our algorithm is based on CalQL, the most advanced algorithm in offline RL. This algorithm is proposed to facilitate efficient online RL fine-tuning; however, our method does not require online RL fine-tuning. We have found that the calibration mechanism is beneficial in addressing issues related to sparse rewards.

\subsection{Practical Implementations}
In this section, we present the practical implementations of \textbf{CO\_RFT}, which include the model architecture of the VLA models employed, as well as a technique known as \textit{Reward Upsampling}. This technique is essential in addressing the challenges associated with sparse rewards.

% \begin{figure}[!htb]
%     \centering
%     \includegraphics[width=1.0\linewidth]{assets/robovlms.pdf}
%     \caption{}
%     \label{fig:robovlms}
% \end{figure}

\paragraph{Model Architecture} We implement our method on RoboVLMs \cite{li2024generalistrobotpoliciesmatters}, which utilize Kosmos2 \cite{peng2023kosmos2groundingmultimodallarge} as the VLM backbone. RoboVLMs exploit historical information using LSTM or GPT architectures; such designs are rarely considered by other VLA models. Following the implementation of RoboVLMs, we developed our methods based on the TD3 \cite{fujimoto2018addressingfunctionapproximationerror} algorithm, recognized as the most advanced RL algorithm that produces deterministic actions.

\paragraph{Sparse Reward Considerations} Sparse rewards represent a realistic characteristic inherent to real-world tasks, presenting a significant obstacle to the reinforcement fine-tuning of VLA models. We identify that the inefficiency of reinforcement learning in the context of sparse rewards partially stems from the inaccurate prediction of the sparse reward signal. 
% To address this issue, we introduce two strategies: \textit{Reward Upsampling} and \textit{Return-to-Go Regularizer}. 
To address this issue, we introduce \textit{Reward Upsampling}, a straightforward data collection strategy that records additional successful steps while gathering demonstrations through human teleoperation. This approach enables us to acquire more samples containing reward signals, thereby mitigating data sparsity in value learning.
% The \textit{Return-to-Go Regularizer} represents the practical implementation of the CalQL regularizer. We emphasize that the Return-to-Go value can serve as a calibration mechanism to regularize the Q-function to a calibrated distribution, thereby facilitating the learning of the Q-function in sparse reward settings.
% \begin{equation}
% \label{eq:calql}
% V^\mu(s) = Return\_to\_Go(s)
% \end{equation}

\section{Experiments}
We evaluate our \textbf{CO-RFT} algorithm on real-world tasks by designing several \textit{grasp and pick} tasks for a dexterous hand. For each task, the robot receives a language instruction that requires common-sense and physical knowledge to identify suitable grasping targets. Subsequently, it executes actions based on visual input and proprietary information. All tasks are performed on a white tabletop under uniform lighting conditions. Through our experiments, we aim to investigate the following questions:

\begin{itemize}
\item How does Chunked RL perform in comparison to standard RL and IL?
\item Can Chunked RL improve sample efficiency and generalization for VLA models?
\item How does data diversity influence the performance of RL in fine-tuning VLA models?
\end{itemize}

\subsection{Experimental Setup}

\begin{figure}[!htb]
    \centering
    \includegraphics[width=1.0\linewidth]{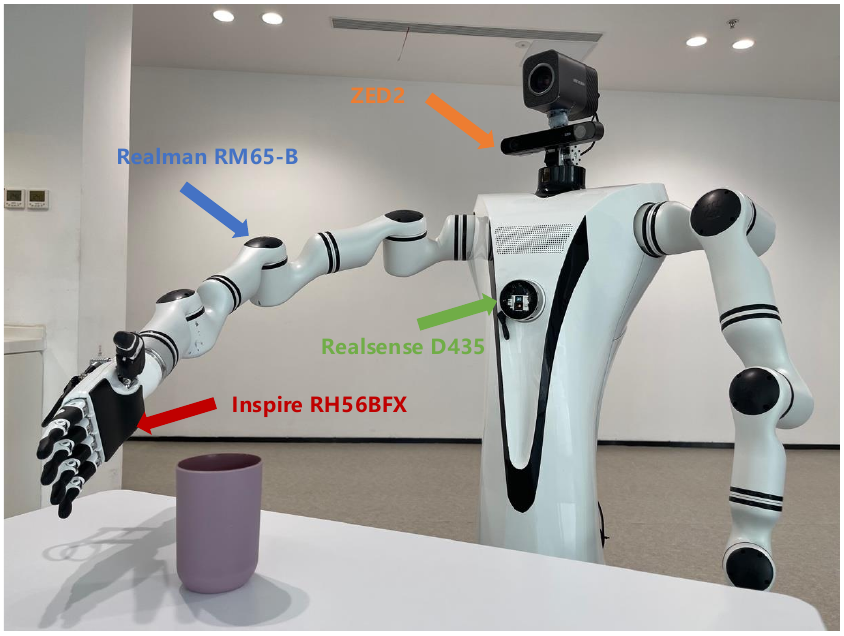}
    \caption{The robot used in our experiment.}
    \label{fig:platform}
\end{figure}

\paragraph{Hardware}
Our experiments are conducted on the Realman Single-arm platform as shown in Figure. \ref{fig:platform}, equipped with a 6 DoF robotic arm. Dexterous robot hands form Inspire Company which has 6 active joint are installed as the end effectors. The stereo camera ZED2 which has 110°(H) × 70°(V) FoV is installed on the head to capture RGB images.

\begin{figure*}[!htb]
    \centering
    % 两个子图水平排列（宽度总和略小于1.textwidth以留间隙）
    \begin{subfigure}{0.3\textwidth}
        \includegraphics[width=\textwidth]{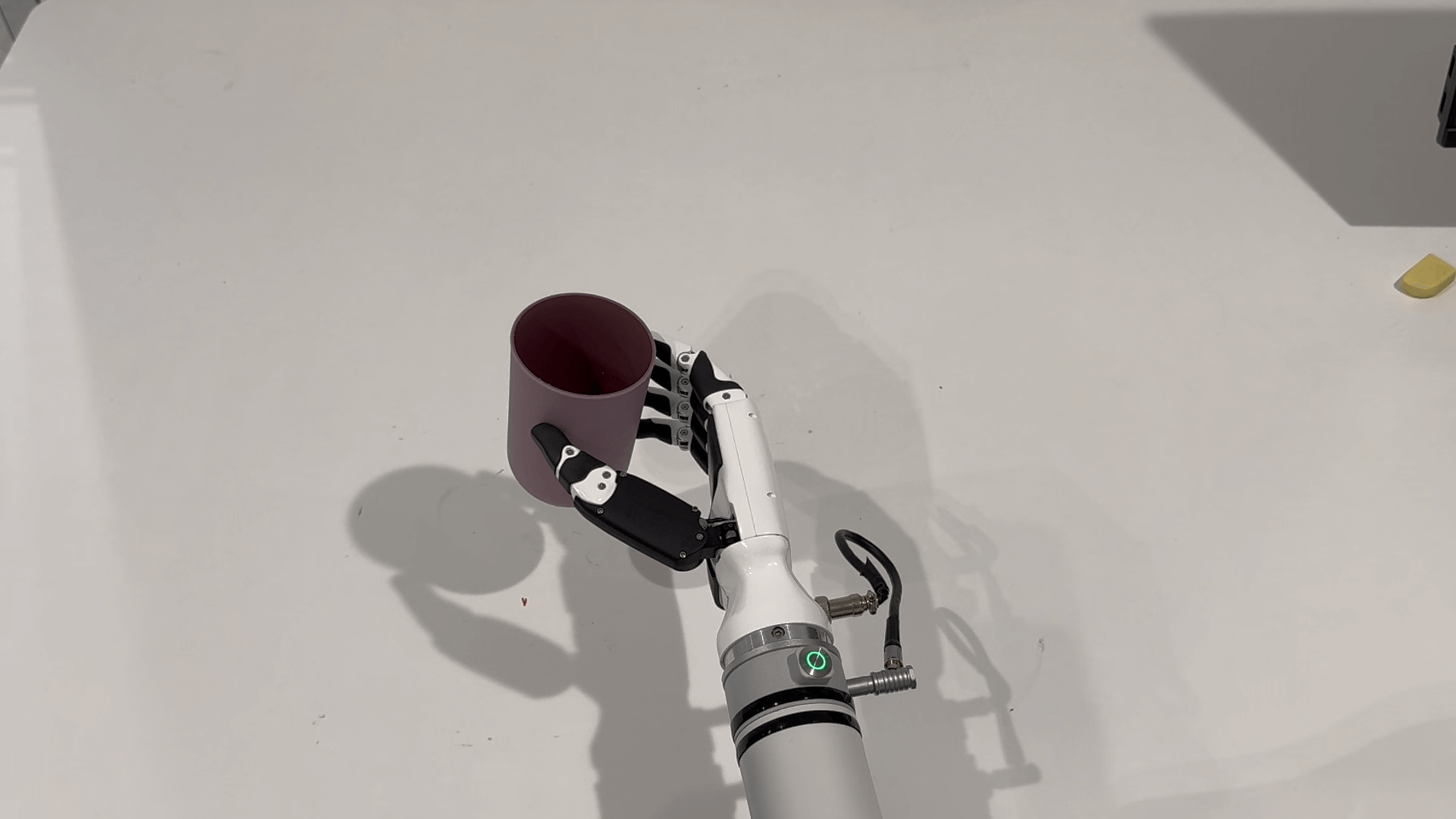}
        \caption{Grasping a cup}  % 子图标题1
        \label{fig:ct}
    \end{subfigure}
    % \hfill % 最大化水平间距
    \begin{subfigure}{0.3\textwidth}
        \includegraphics[width=\textwidth]{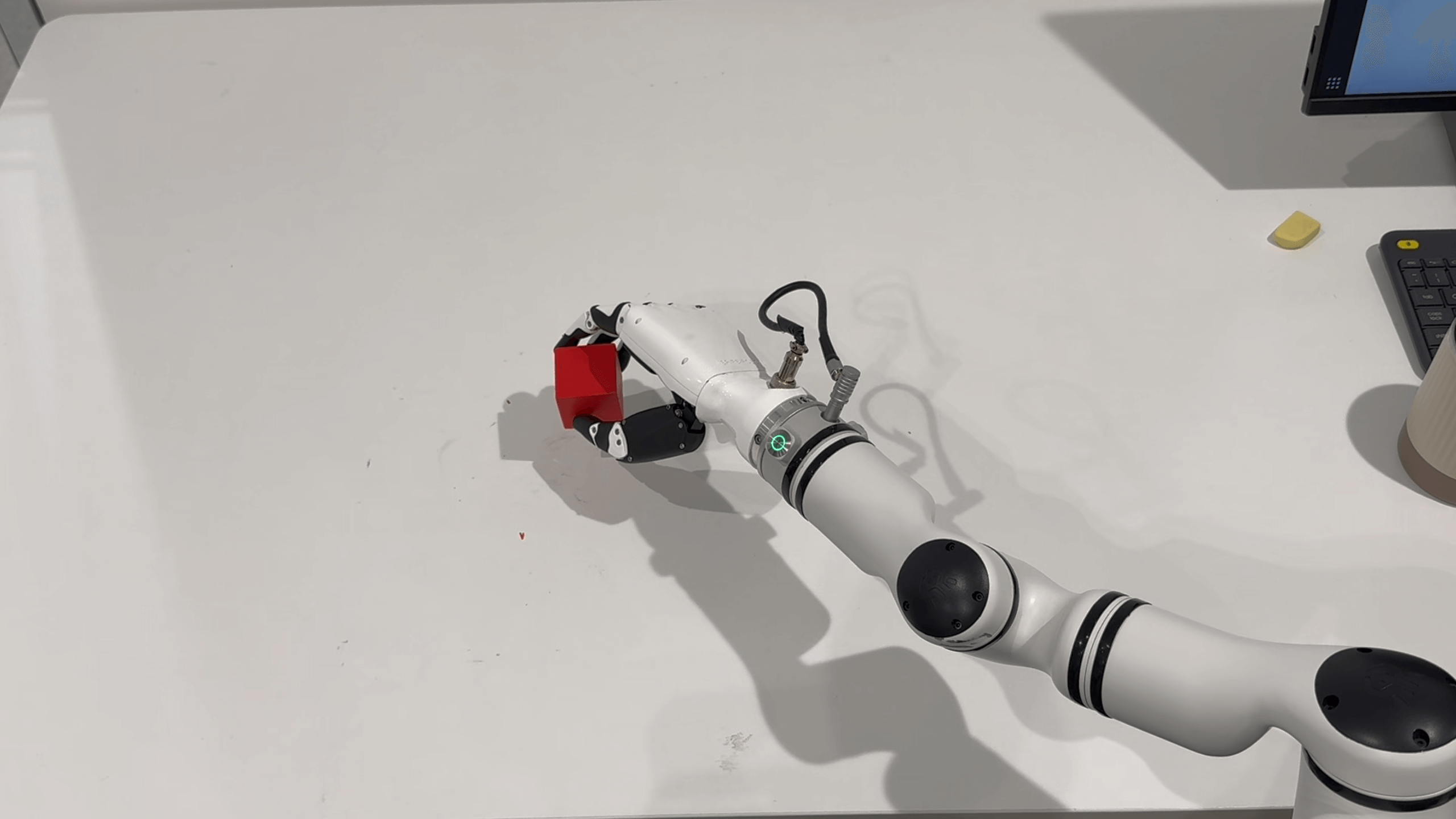}
        \caption{Pinching a cube}  % 子图标题2
        \label{fig:radar}
    \end{subfigure}
    \begin{subfigure}{0.3\textwidth}
        \includegraphics[width=\textwidth]{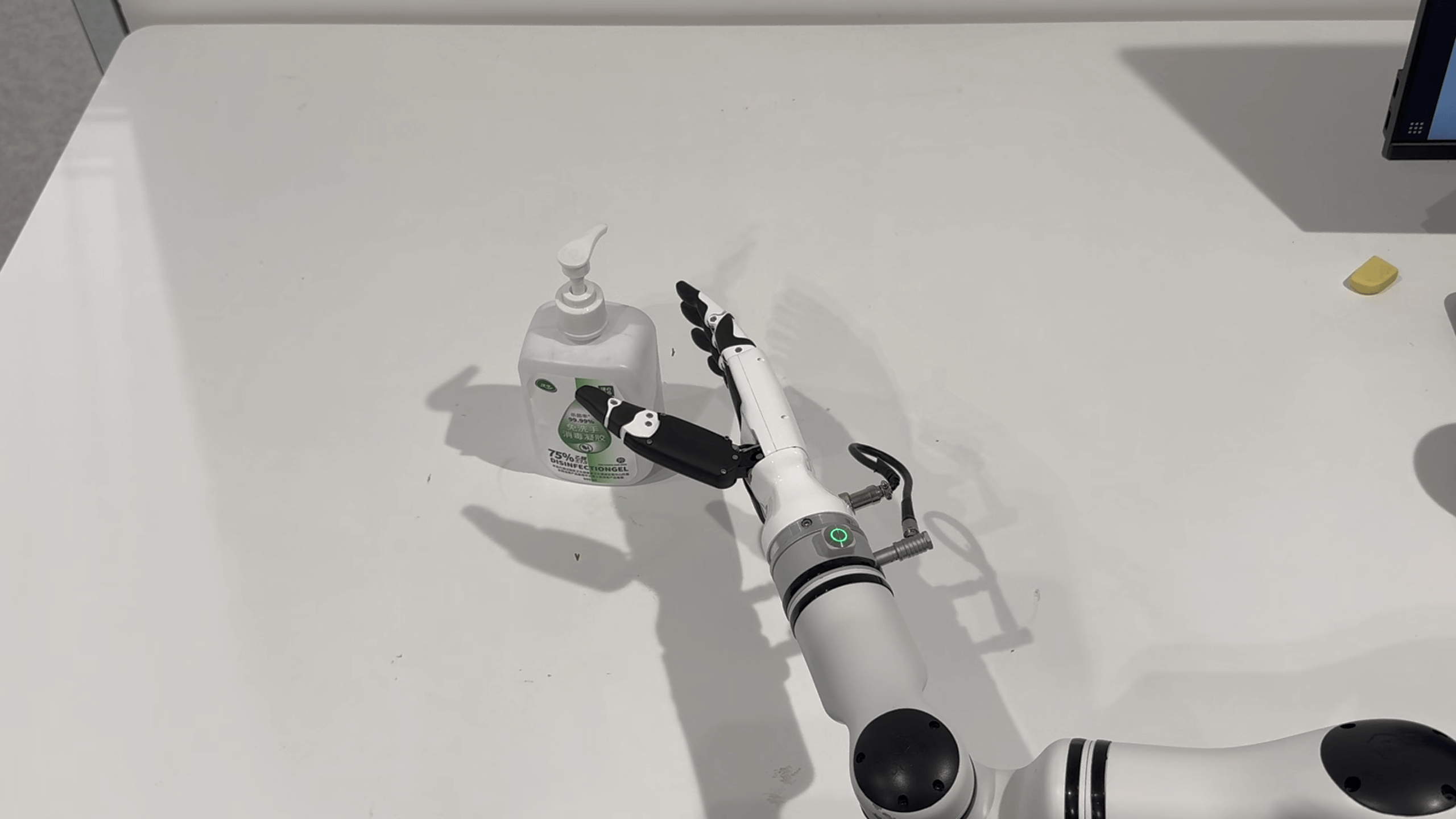}
        \caption{Grasping a sanitizer}  % 子图标题2
        \label{fig:radar}
    \end{subfigure}
    \begin{subfigure}{0.3\textwidth}
        \includegraphics[width=\textwidth]{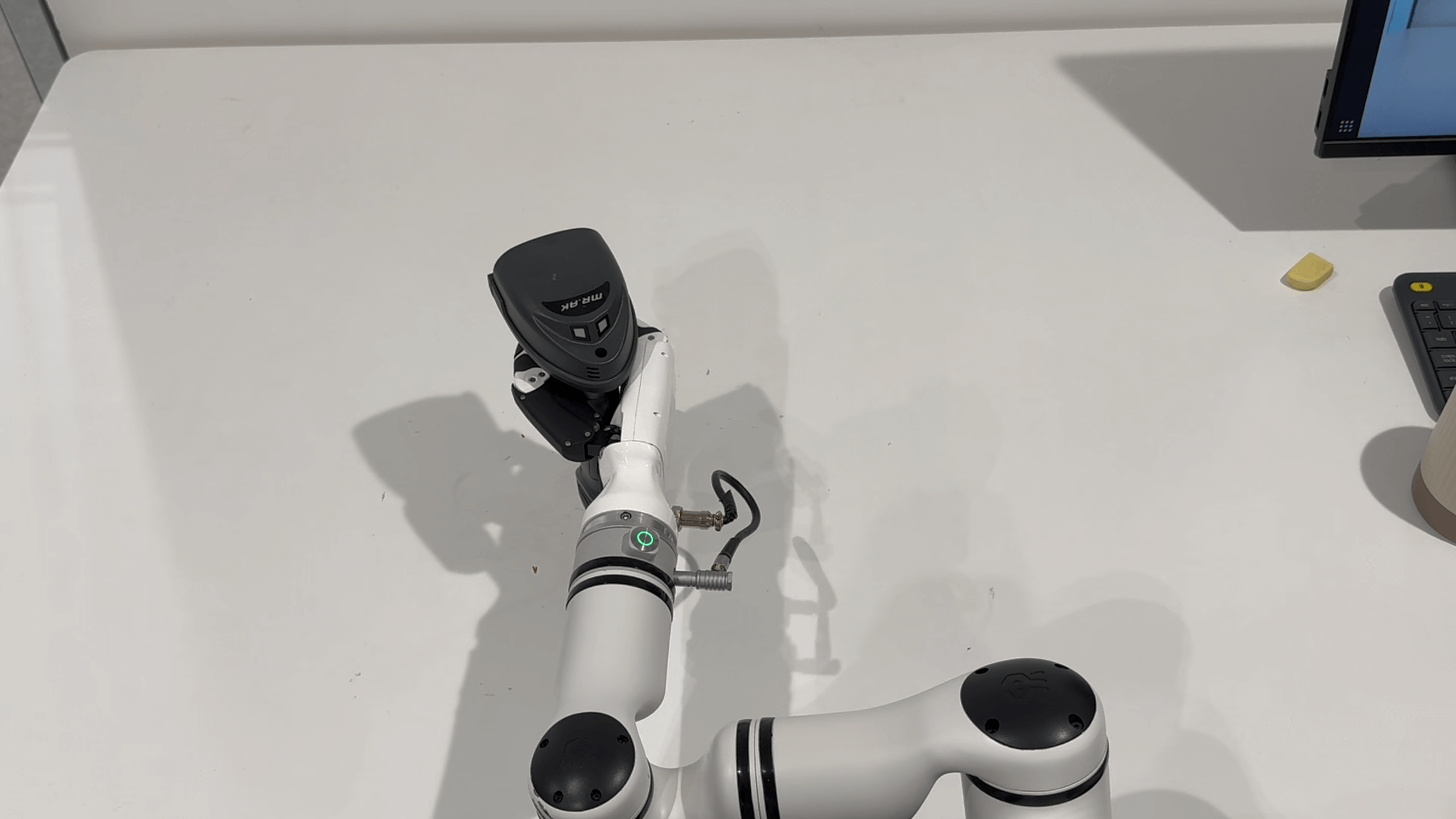}
        \caption{Holding a barcode scanner}  % 子图标题1
        \label{fig:ct}
    \end{subfigure}
    % \hfill % 最大化水平间距
    \begin{subfigure}{0.3\textwidth}
        \includegraphics[width=\textwidth]{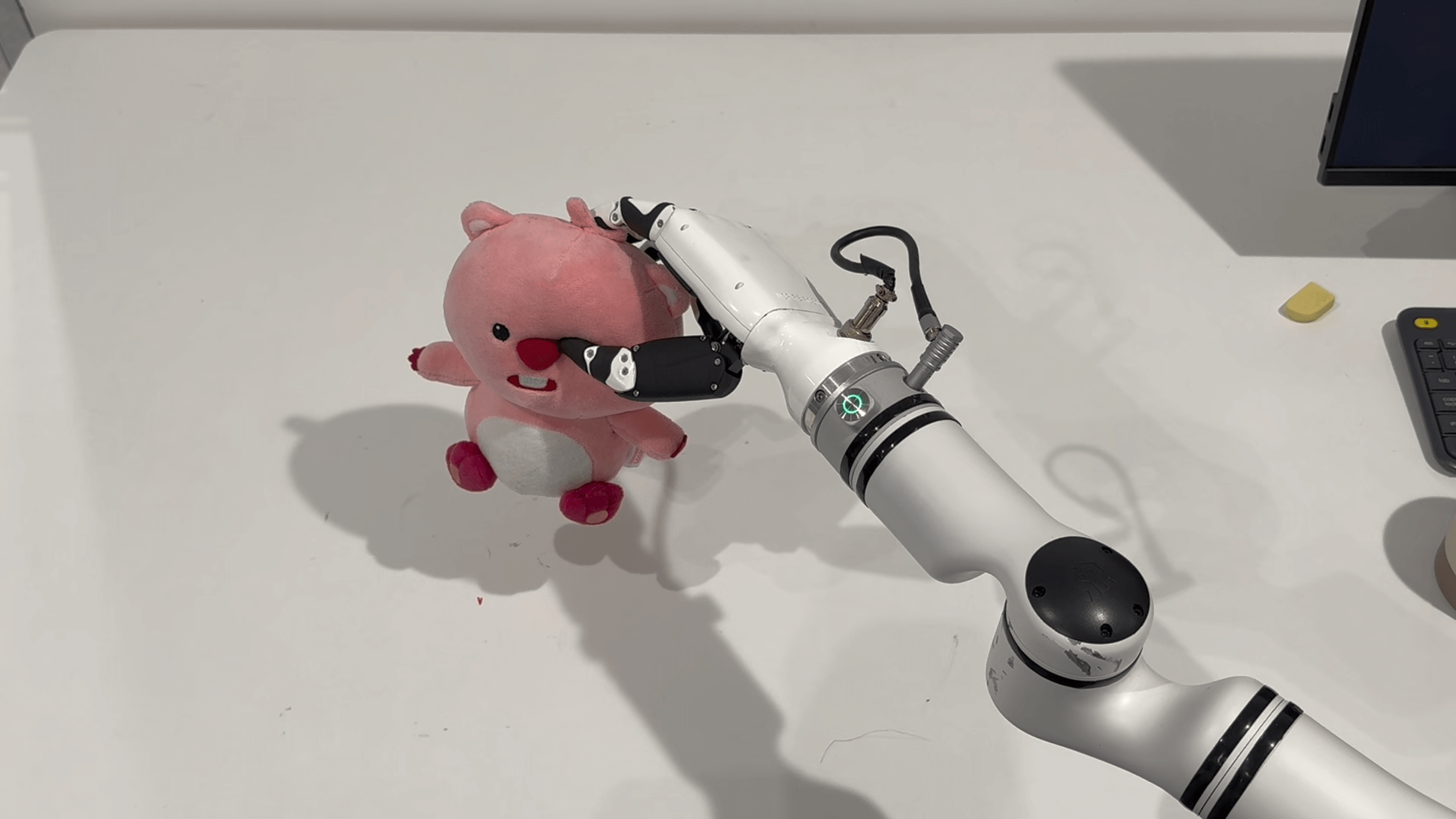}
        \caption{Catching a loop}  % 子图标题2
        \label{fig:radar}
    \end{subfigure}
    \begin{subfigure}{0.3\textwidth}
        \includegraphics[width=\textwidth]{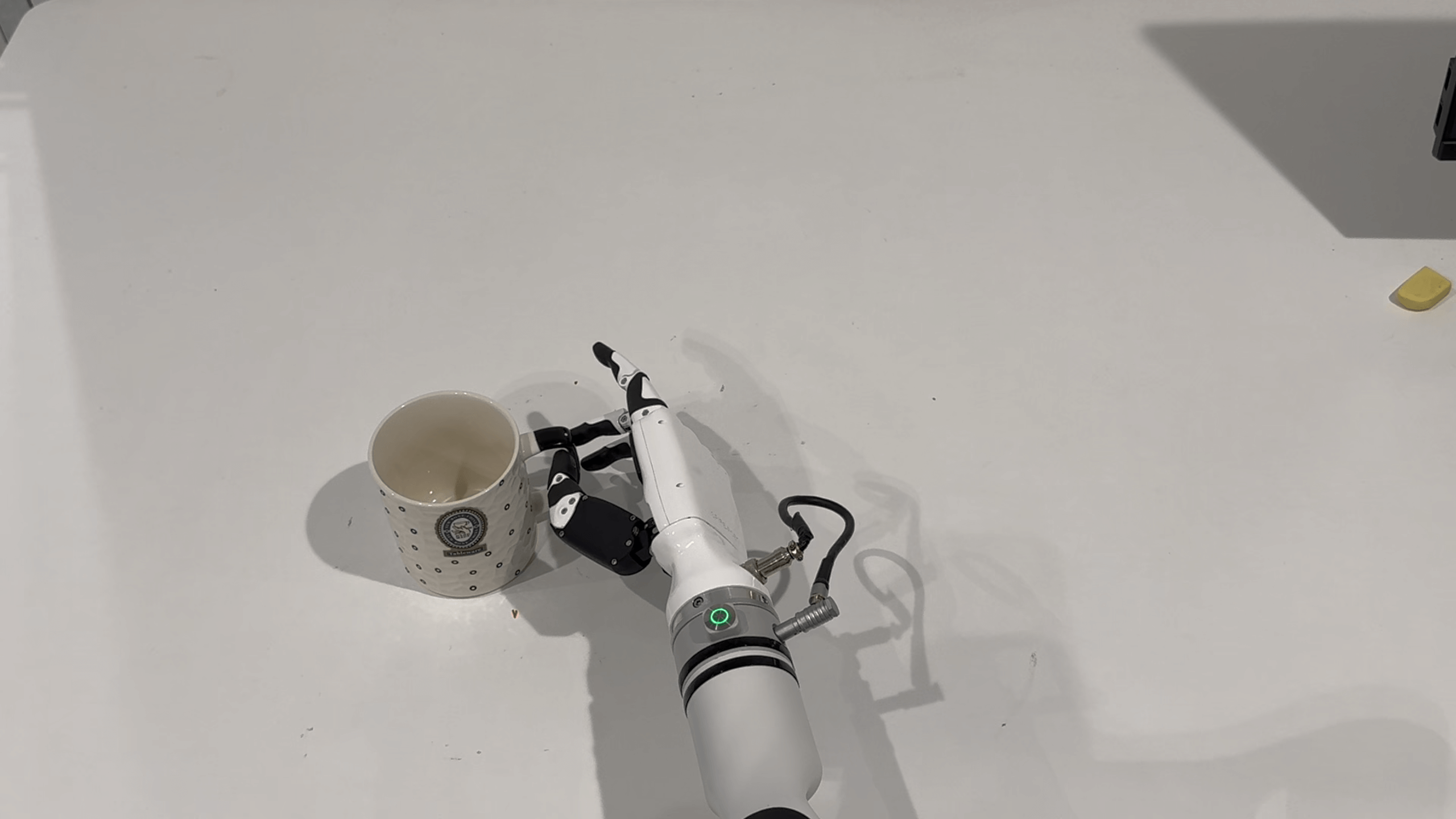}
        \caption{Retrieving a mug}  % 子图标题2
        \label{fig:radar}
    \end{subfigure}
    % 主图标题（描述整体）
    \caption{Visualization of the real-world tasks.}
    \label{fig:robot_combo}
\end{figure*}

\paragraph{Tasks and Datasets}
In our experiments, we designed 6 challenging dexterous manipulation tasks to evaluate the effectiveness of the proposed method. 
% The single-hand tasks include grasping a cup, retrieving a mug, holding a barcode scanner, catching a loop, pinching a cube, and grasping a sanitizer.
For the tasks described in Table \ref{fig:robot_combo}, we need to collect human demonstration data for learning. Specifically, we used strain gauge gloves and a positioning camera as teleoperation devices to control a dexterous hand for data collection. During the collection of demonstration data for each task, the initial position of the dexterous hand is fixed. the position of the objects is not fixed but is randomly placed within a rectangular area.

% \begin{figure*}[htb]
%     \centering
%     \includegraphics[width=1.0\linewidth]{assets/tasks_all.pdf}
%     \caption{Illustration of the robot performing each task with our method. For each task, the robot starts from a predefined initial pose and subsequently performs the task by utilizing the proprioceptive state and image observations. The process continues until the robot receives a reward for successfully completing the task, as indicated within the green box.}
%     \label{fig:tasks}
% \end{figure*}

% \begin{table*}[!h]
% \centering
% \small  % Reduce font size for double-column format
% \scalebox{1.2}{
% \begin{tabular}{@{}c c c c c c@{}}
% \toprule
% \textbf{Tasks} & \textbf{Demos} & \textbf{Image Inputs} & \textbf{Reset} & \textbf{Training Time} & \textbf{Classifier} \\
% \midrule
% Cup Taking      & 40 & Chest Camera, Head Camera & Fixed & 59 mins & 98\% \\
% Cube Taking     & 36 & Chest Camera, Head Camera & Fixed & 72 mins & 96\% \\
% Scanner Taking  & 31 & Chest Camera, Head Camera & Random & 41 mins & 95\% \\
% Loopy Taking    & 36 & Chest Camera, Head Camera & Random & 42 mins & 99\% \\
% \bottomrule
% \end{tabular}
% }
% \vspace{0.5em}
% \captionsetup{labelfont=bf, skip=6pt}
% \caption{Task Parameters. During demo collection for both BC and RL, as well as online training, each episode’s initial end-effector pose resets uniformly at random within a fixed region}
% \label{tb:task_parameters}
% \end{table*}

\paragraph{Metrics} In the experiment, the robot receives two images from the head camera and grasps the object based on natural language commands. It is considered successful when the object is lifted into the air by the dexterous hand. The model's performance is evaluated using the success rate (SR) and the average cycle time (CT).

\subsection{Experimental Results}

\begin{figure*}[!htb]
    \centering
    % 两个子图水平排列（宽度总和略小于1.textwidth以留间隙）
    \begin{subfigure}{0.65\textwidth}
        \includegraphics[width=\textwidth]{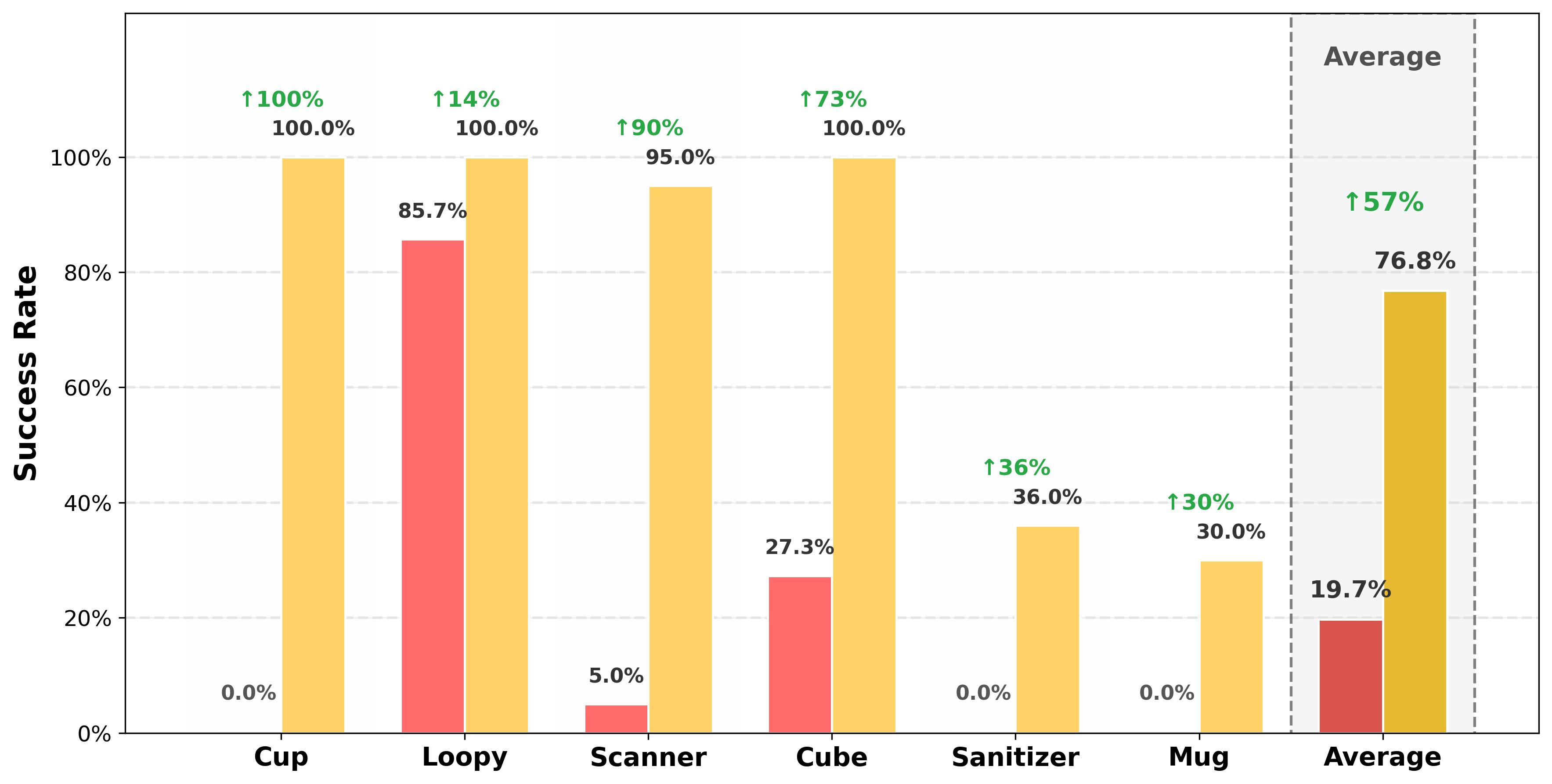}
        \caption{Success rate comparisons in in-distribution (IND) scenarios.}  % 子图标题1
        \label{fig:sr_ind}
    \end{subfigure}
    % \hfill % 最大化水平间距
    \begin{subfigure}{0.32\textwidth}
        \includegraphics[width=\textwidth]{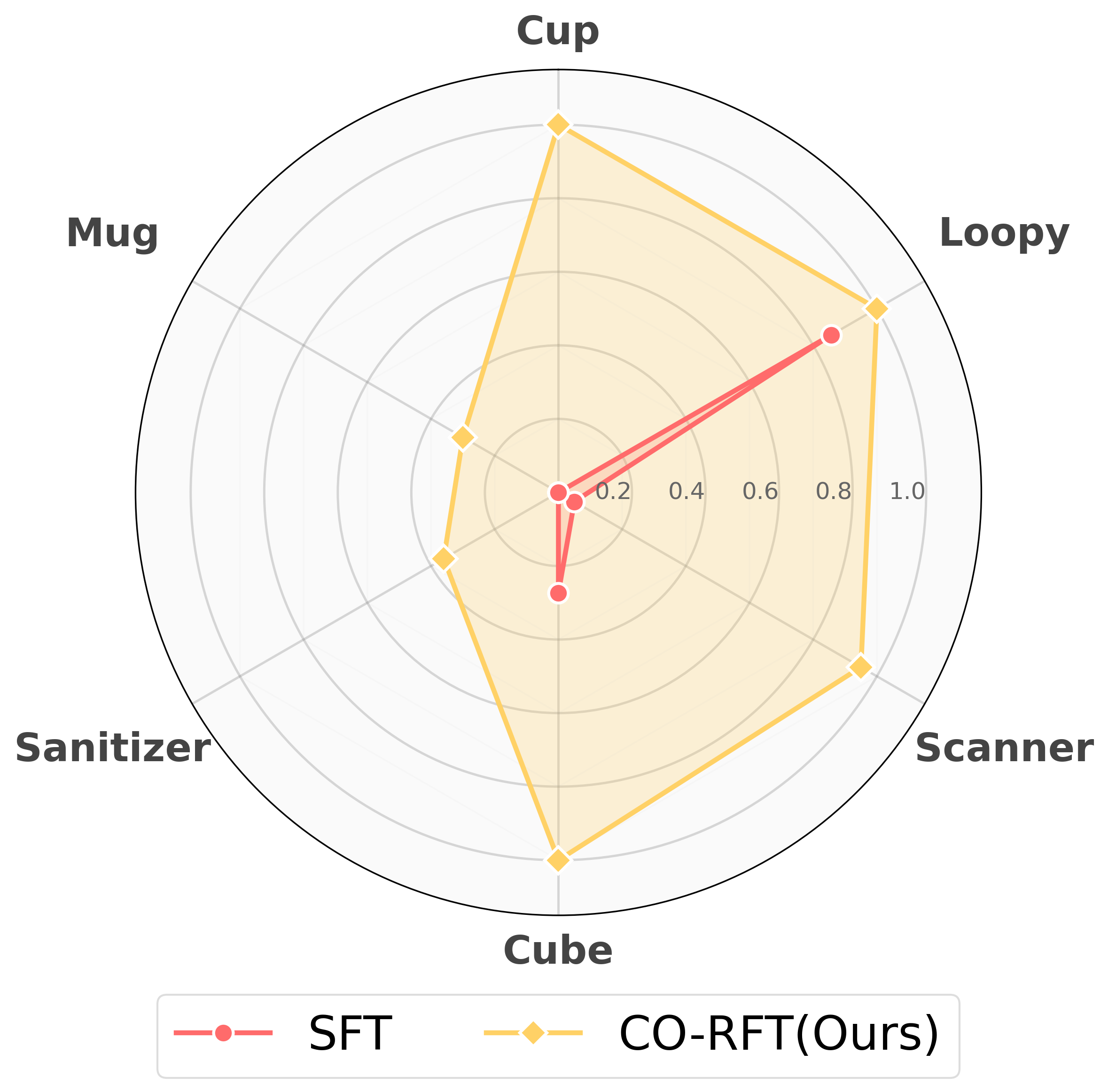}
        \caption{Radar chart of success rate.}  % 子图标题2
        \label{fig:radar}
    \end{subfigure}
    % 主图标题（描述整体）
    \caption{Success rate comparisons between SFT and CO-RFT: In evaluations comprising 40 trials per task, learned CO-RFT policies exhibited a substantial advantage over BC policies, with an average improvement of 57\% in IND scenarios.}
    \label{fig:ind_comparison}
\end{figure*}

\begin{figure}[!htb]
    \centering
    \includegraphics[width=1.0\linewidth]{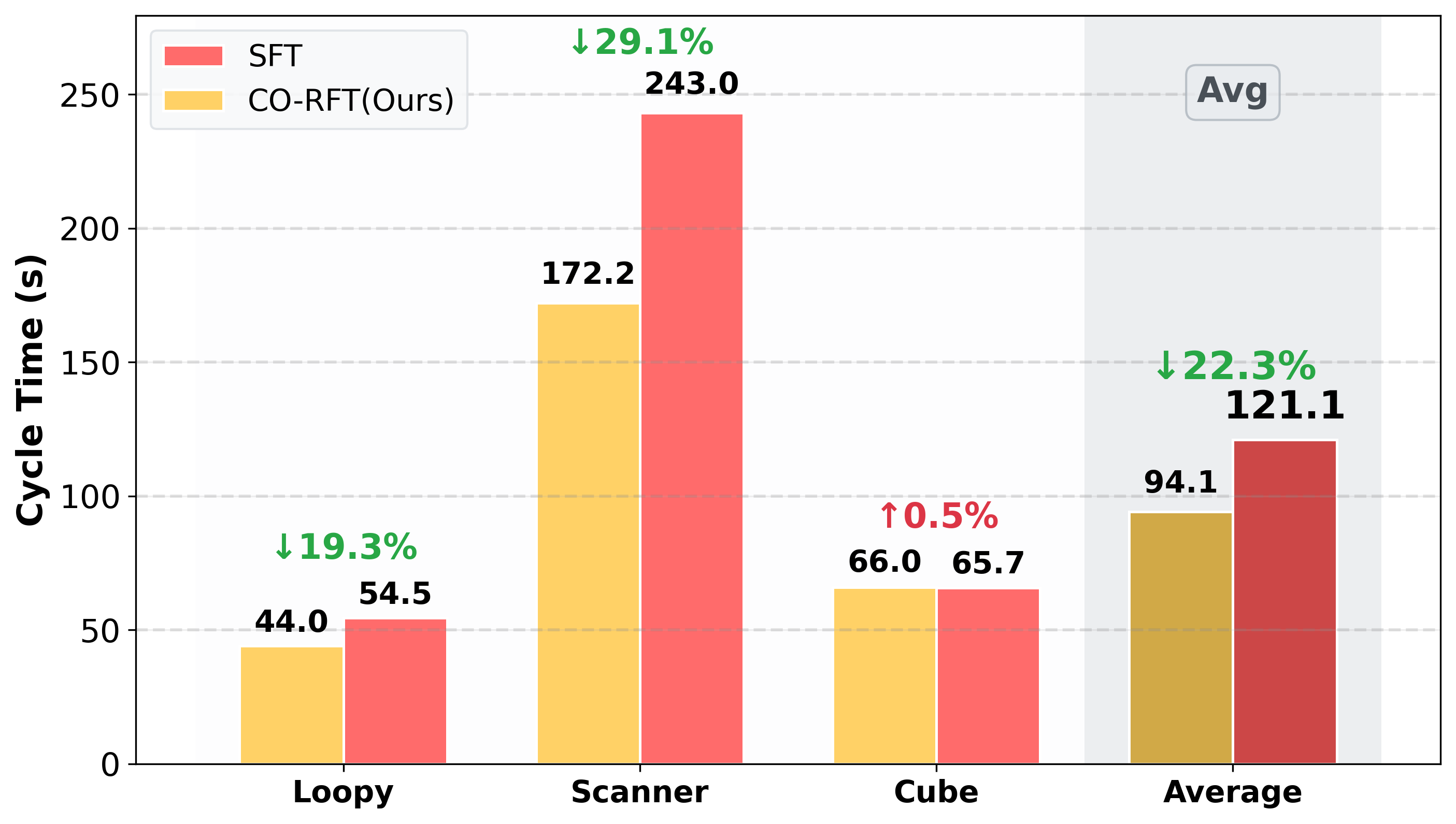}
    \caption{Cycle time comparisons: We recorded the average steps taken for the robot to succeed in each task.  CO-RFT policies outperformed SFT policies, demonstrating an average reduction of 22.3\% in cycle time. It is noteworthy that cycle times for the remaining tasks are excluded due to the consistent failures of SFT policies on these tasks.}
    \label{fig:ct_comparison}
\end{figure}

\paragraph{Results}
As illustrated in Figures \ref{fig:ind_comparison} and \ref{fig:ct_comparison}, CO-RFT significantly outperforms the SFT method, achieving an average improvement of 57\% in success rate and an average reduction of 22.3\% in cycle time. This approach effectively masters several tasks, attaining a nearly 100\% success rate in four out of six tasks. Notably, in the tasks of grasping a cup, grasping a sanitizer, and retrieving a mug—tasks in which the SFT method consistently fails, CO-RFT demonstrates promising performance.
From Figure \ref{fig:sr_ind}, we observe that the SFT method fails on most tasks. This failure is attributed to the insufficient demonstrations provided for VLA models to fully learn these tasks. Evidence suggests that more than 100 samples are necessary for VLA models to effectively acquire a novel skill using SFT; however, we only have 30 samples available. In contrast, CO-RFT demonstrates significantly improved performance, effectively utilizing the available samples to achieve higher success rates across the evaluated tasks. This indicates that CO-RFT achieves much better sample efficiency by employing offline RL and Chunked RL. 
However, we observe that both SFT and CO-RFT exhibit low success rates in the tasks of grasping a sanitizer and retrieving a mug. The challenges associated with the grasping a sanitizer task are primarily due to the sanitizer's color being particularly similar to the white background, which complicates the model's ability to accurately identify the key grasping points. In the case of the retrieving a mug task, the model must grasp the handle of the cup, requiring fine-grained manipulation that is inherently difficult to master. Despite these challenges, CO-RFT achieves a success rate of 36\% in the grasping a sanitizer task and 30\% in the retrieving a mug task, demonstrating the potential to distinguish subtle visual features and develop fine-grained manipulation skills.

\paragraph{Cycle Time Comparisons}
As illustrated in Figure \ref{fig:ct_comparison}, CO-RFT results in an average reduction of 22.3\% in cycle time. This indicates that CO-RFT, by incorporating RL, has the potential to surpass expert demonstrations and achieve a more efficient policy. However, CO-RFT fails to demonstrate any improvement in cycle time for the pinching a cube task. Since the demonstrations for this task are initialized from a fixed position, the lack of data diversity hinders the ability of Offline RL to learn an expressive value function, which is essential for enhancing the policy.

\begin{figure}[!htb]
    \centering
    \includegraphics[width=1.0\linewidth]{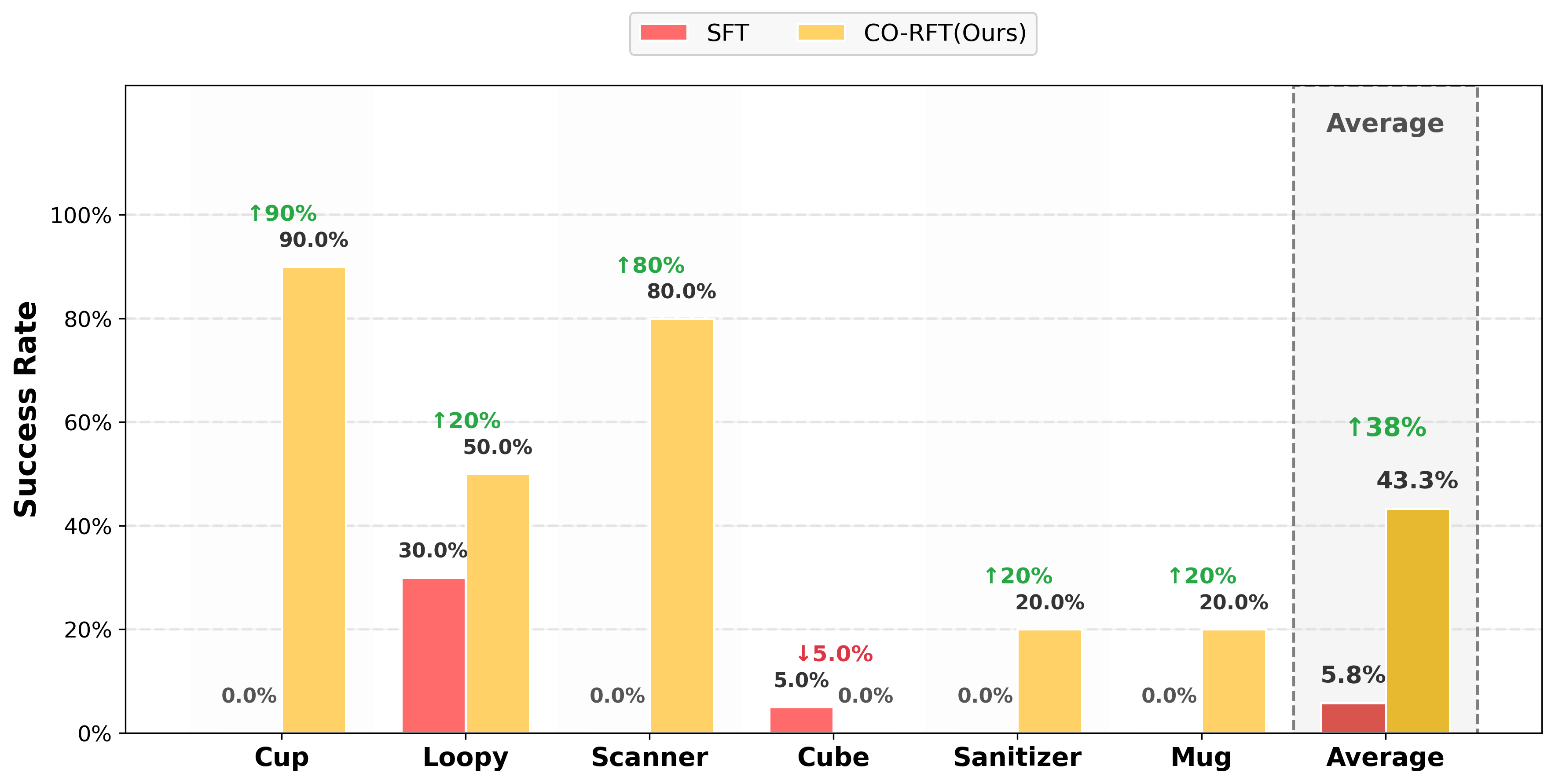}
    \caption{Success rate comparisons in OOD scenarios: For each task, the target object is initialized in unseen positions, and the corresponding success rates are recorded.}
    \label{fig:ood}
\end{figure}

\paragraph{Positional Generalization Analysis}
As illustrated in Figure \ref{fig:ood}, CO-RFT significantly outperforms SFT in out-of-distribution (OOD) scenarios. Specifically, CO-RFT achieves a 38\% improvement in success rate over SFT and demonstrates superior performance across most tasks. Notably, in the tasks of grasping a cup, catching a loop, and holding a barcode scanner, CO-RFT attains success rates of 90\%, 50\%, and 80\%, respectively, showcasing impressive positional generalization capabilities. In contrast, when trained with identical datasets, SFT fails to perform adequately on the majority of tasks when deployed in unfamiliar positions. This evidence implies that offline RL, despite lacking the ability to learn from online exploration, can still achieve remarkable generalization capabilities. 
% We attribute this ability to Q-learning's focus on evaluating states and actions while prioritizing critical decisions.

\begin{table}[!htb]
\centering
\resizebox{0.48\textwidth}{!}{
\begin{tabular}{lcccc}
\toprule
Initialization & {Task} & \multicolumn{1}{c}{IND} & \multicolumn{1}{c}{OOD} & \multicolumn{1}{c}{Performance Drop} \\
\midrule
\multirow{4}{*}{Random} &  Cup       & 100\%      & 90\%     & -10\%  \\
 & Scanner   & 95\%   & 80\%    & -15\%  \\
 & Mug       & 30\%    & 20\%   & -10\%  \\
 & \textbf{Average}  & 75\%     & \textbf{63.3\%} & \textbf{ -11.7\%} \\

\midrule

\multirow{4}{*}{Fixed} &  Cube & 100\%    & 0\%     &  -100\%  \\
 &  Loopy     & 100\%  & 50\%    &  -50\%  \\
 &  Sanitizer & 36\%  & 20\%     &  -16\%   \\
 &  \textbf{Average}  & \textbf{78.7\%}  & 23.3\%  & 55.3\%  \\
\bottomrule
\end{tabular}
}
\caption{Generalization ability comparison: We collect two types of demonstrations. One involves grasping an object in a fixed position, referred to as fixed-initialized datasets, while the other entails grasping an object that is randomly placed within a rectangular area, termed random-initialized datasets. We train CO-RFT on these distinct datasets and record the IND and OOD performance.}
\label{tab:data_diversity}
\vspace{0.1in}
\end{table}

\paragraph{Importance of Data Diversity}
As depicted in Table \ref{tab:data_diversity}, data diversity plays a crucial role in the generalization ability of CO-RFT. In IND scenarios, CO-RFT consistently achieves a high success rate, regardless of the dataset collection methods employed. This indicates that CO-RFT can effectively master these IND scenarios.
However, an examination of the OOD performance of CO-RFT trained on different datasets reveals significant variability. Specifically, models trained on randomly initialized datasets exhibit only minor performance degradation, ranging from 10\% to 15\%. In contrast, those trained on fixed-initialized datasets experience catastrophic performance degradation, averaging 55.3\%.
This phenomenon aligns with the principles of RL, wherein a more comprehensive dataset that encompasses a greater variety of states and actions leads to a more accurate value function. Consequently, this enhances both the performance and generalization of the policy.

\section{Conclusion}
In this work, we propose \textbf{CO-RFT}, a two-stage reinforcement learning algorithm designed to achieve sample-efficient reinforced fine-tuning of VLA models using a limited set of demonstrations. Our algorithm employs a method termed \textbf{Chunk RL}, which integrates action chunking into the reinforced fine-tuning process of VLA models, resulting in enhanced sample efficiency and notable positional generalization capabilities. Furthermore, \textbf{CO-RFT} successfully transfers the Robovlms, trained with gripper embodiment, to dexterous hand embodiment, mastering several real-world tasks with only 30 to 60 samples.
Our empirical results demonstrate that Offline RL can surpass IL in terms of success rate, cycle time, and generalization ability using the same datasets, indicating that Offline RL has the potential to serve as a superior paradigm for fine-tuning VLA models on offline datasets.
While promising, our approach still has limitations. For instance, the proposed methods encounter over-training issues and require additional techniques to validate the optimal checkpoint. Meanwhile, we adopt a deterministic policy as our action head, which may fail to accurately model the multi-modal action distribution. Future work will involve training diffusion-based policies in conjunction with reinforcement learning.

\bibliography{aaai2026}

\end{document}